\documentclass{article} 
\usepackage{iclr2023_conference_tinypaper,times}
\usepackage{comment}
\usepackage{adjustbox}

\usepackage{amsmath,amsfonts,bm}

\def\eqref#1{equation~\ref{#1}}

\def\1{\bm{1}}

\DeclareMathAlphabet{\mathsfit}{\encodingdefault}{\sfdefault}{m}{sl}
\SetMathAlphabet{\mathsfit}{bold}{\encodingdefault}{\sfdefault}{bx}{n}

\usepackage{hyperref}
\usepackage{url}
\usepackage{graphicx, pdflscape, pdfpages, multirow}
\usepackage[normalem]{ulem}

\title{Observations on Building RAG Systems for Technical Documents}

\author{Sumit Soman and Sujoy Roychowdhury\thanks{Global AI Accelerator,
Ericsson R\&D, Bangalore, India. Both authors contributed equally. \href{https://anonymous.4open.science/r/RAG_ICLR-55CB/README.md}{Git Repo Link.}} \\
\texttt{\{sumit.soman, sujoy.roychowdhury\}@ericsson.com} \\
}

\iclrfinalcopy 
\begin{document}

\maketitle

\begin{abstract}

Retrieval augmented generation (RAG) for technical documents creates challenges as embeddings do not often capture domain information. We review prior art for important factors affecting RAG and perform experiments to highlight best practices and potential challenges to build RAG systems for technical documents.
\end{abstract}

\section{Introduction}

Long form Question Answering (QA) involves generating paragraph-size responses from Large Language Models (LLMs). {\color{black} RAG for technical documents has several challenges \cite{xu2023retrieval, toro2023dynamic}}. Factors affecting retrieval performance, including in-context documents, LLMs and metrics, have been evaluated \cite{chen2023understanding}. To further build on this work, we conduct experiments on technical documents with {\color{black}telecom and battery terminology} to examine the influence of chunk length, keyword-based search and ranks (sequence) of retrieved results in the RAG pipeline. 


\section{Experimental Setup}
Our experiments are based on IEEE Wireless LAN Medium Access Control (MAC) and Physical Layer (PHY) specifications \cite{9363693} {\color{black}and IEEE Standard Glossary of Stationary Battery Terminology \cite{7552407}}. We separately process the glossary of definitions and the full document, as many expected questions are based on the definitions.{\color{black} We source questions based on domain knowledge and report experimental results on 42 representative queries across the documents. }
{\color{black} Multiple embedding models can be used, \cite{reimers2019sentence}}, we use MPNET \cite{song2020mpnet} for the entire document - excluding tables and captions. For the glossary, we split the term and the definition and generate separate embeddings for them, as well as for the full paragraph having the defined term and the definition. \cite{soman2023observations} have reviewed other LLMs for telecom domain, but we chose llama2-7b-chat model \cite{touvron2023llama} as it is free and has a commercial-friendly license.  
We evaluate on multiple questions and report on selected questions to substantiate our observations. {\color{black}For reference, the prompts used for the LLM are provided in Appendix \ref{app:appA}.}



\section{Observations}
\begin{figure}[h]
    \centering
    \includegraphics[scale=0.32]{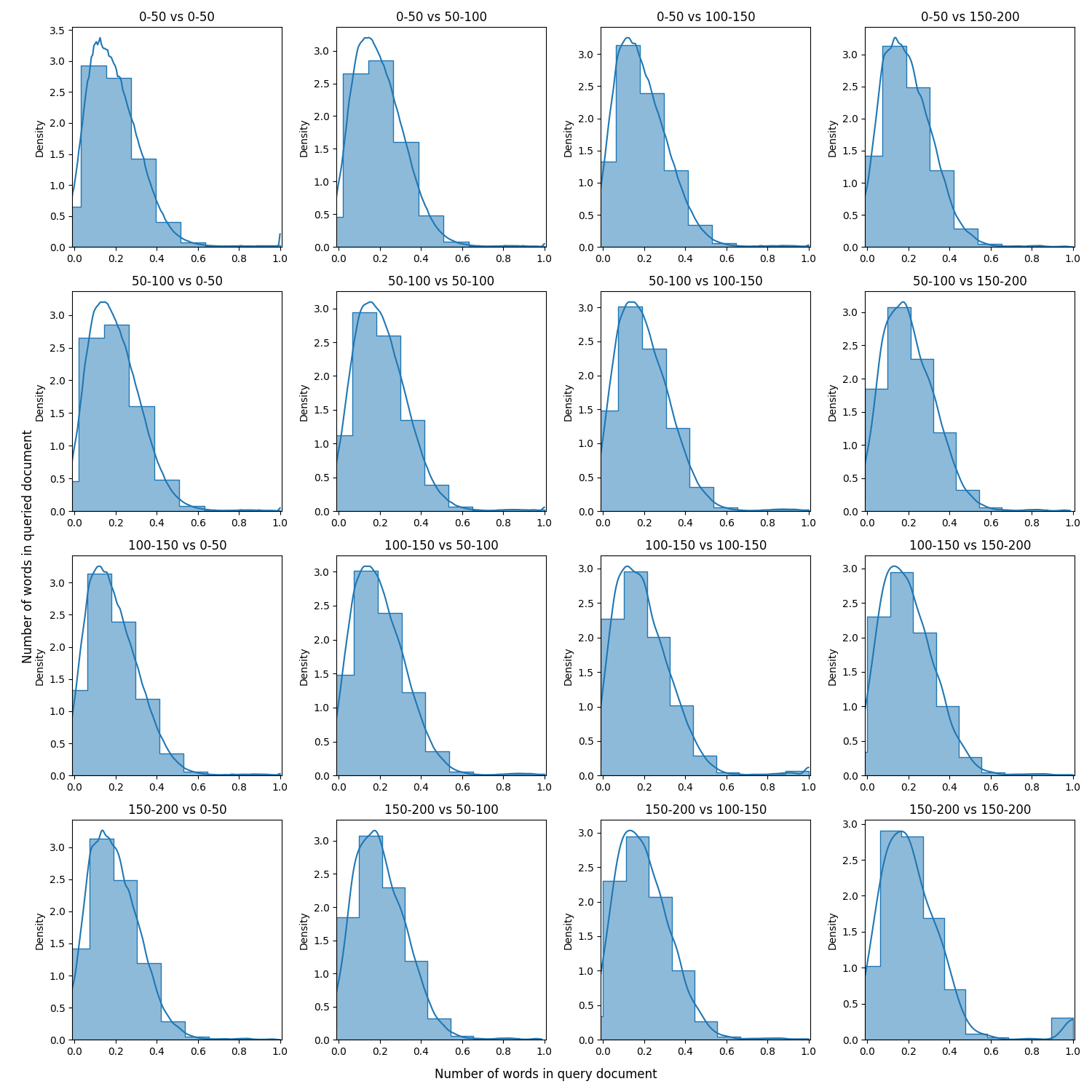}
    \caption{The distribution of similarities across 10974 documents of various sizes split by number of words in the document}
    \label{fig:kde_chunk}
\end{figure}

We first observe that sentence embeddings become unreliable with increasing chunk size. Appendix \ref{app:appB} Fig. \ref{fig:kde_chunk} shows the Kernel Density Estimate (KDE) plot of cosine similarity scores for various sentence lengths. We take 10,970 sentences and look at pairwise similarity for all the sentences. A high similarity is observed when the length of the sentences is relatively long. The higher similarity distribution for larger lengths indicates spurious similarities which we manually validate for a few samples. We find that when both the query and queried document are over 200 words, the similarity distribution is bimodal. When either of them are over 200 words, there is a small but less perceptible lift at higher similarities. 

\begin{table}[h]
\centering
\scalebox{0.8}{
\begin{tabular}{|p{0.8cm}|p{3.8cm}|p{5.7cm}|p{1.8cm}|}
\hline
\multicolumn{1}{|c|}{\textbf{Hyp}} & \multicolumn{1}{c|}{\textbf{Hypothesis}}                                  & \multicolumn{1}{c|}{\textbf{Observation}}                                                                                                      & \multicolumn{1}{c|}{\textbf{Support (Samples)}} \\ \hline
H1                                    & Splitting definition and defined words help in queries                    & For definitions, using the defined word and definition separately for retrieval gives better performance                                       & 22 of 30 queries (ID 2, 3)                             \\ \hline
H2                                    & Similarity scores should not be  used to compare retrieved results                     & We observe that similarity scores between different approaches are not comparable and absolute values are often very small for correct answers & 24 of 30 queries (ID 2, 3)                             \\ \hline
\multirow{2}{*}{H3}                   & \multirow{2}{*}{Position of keywords matter}                              & Keywords closer to the beginning of the sentence are retrieved with high accuracy                                                              & \multirow{2}{*}{25 of 30 queries }                           \\ \cline{3-3} 
                                      &                                                                           & Keywords which occur later in the sentence are difficult to be retrieved                                        &   (ID 1, 4, 5, 6)
                                      \\ \hline
H4                                    & Sentence Based Similarity is better                                       & Similarity based on sentence and distinct paragraphs retrieved gives much detailed context to generator                                        & ID F1 - Table 2   (8 of 10 queries)                      \\ \hline
H5                                    & Generator for sentence based similarity                                   & Generated answer using sentence based similarity and paragraph based retrieval gives better results                                            & 8 of 10 queries (App. Table 3 - ID F1)                         \\ \hline
H6                                    & Definitions with acronyms or words having acronyms don't perform well & Generated answers often expand or provide abbreviations which is not helpful                                                                   & 15 of 16 queries (App. Table 3 - ID F2, F3)                             \\ \hline
H7                                    & Order of retrieved paragraphs in generator results                        & Order of retrieved paragraphs do not affect generator results in our experiments                                                               & NA                                           \\ \hline
\end{tabular}
}
\caption{Summary of observations - details of individual queries in Appendix \ref{app:appB}}
\label{tab:observations}
\end{table}

Table \ref{tab:observations} summarizes our hypotheses and key observations - corresponding sample queries and their results are provided in Appendix \ref{app:appC}. We hypothesize that splitting on definition and terms can help improve results (H1), similarity scores being a good measure (H2), position of keywords influencing results (H3), sentence-based similarity resulting in a better retriever (H4) and generator (H5), answers for definitions based on acronyms  (H6) and effect of order of retrieved results on generator performance (H7). Of these, H2 is a result of our experiments with distributions of similarity scores referred earlier and H7 is based on \cite{chen2023understanding}. Others are derived from our experiments to improve results. {\color{black}For each  hypotheses, we provide the number of experiments that support the claim  and those that are valid for the same in the last column, along with sample queries.}

We find that retrieval by thresholding on similarity scores is not helpful. For queries 1, 2 and 5, when the query phrase is present in the term or definition, top retrieved score is higher. For query 3, the correct result is retrieved at the second position using definition embedding, but in other cases, result is not retrieved and similarity scores are close. For queries 4 and 6, we are unable to retrieve the correct result, though scores indicate otherwise. Thus, thresholding retriever results based on similarity scores can potentially result in sub-optimal generator augmentation. We evaluate generator performance on our queries based on the retrieved results. This is done using the top $k$ retrieved (a) definitions, and (b) terms and definitions. Better context gives better generated responses. For acronyms and their expansions, the generator does not add any additional value.

For retrieval on the full document, we explore similarity search by sentence and paragraph separately. In the former, we retrieve the paragraph to which the sentence belongs and take top-$k$ distinct paragraphs from top similar sentences. We observe that the results by sentence-based similarity search and paragraphs being used for generator provides better retriever and generator performance. Authors in \cite{chen2023understanding} mention order of presented information to be important, but we did not observe different results on permuting the retrieved paragraphs. We observe generator responses to sometimes fail due to incorrect retrieval, hallucinated facts or incorrect synthesis as highlighted in \cite{chen2023understanding}. We recommend such approaches for definition QA and long form QA.

\section{Conclusions and Future Work}
We show that chunk length affects retriever embeddings, and generator augmentation by thresholding retriever results on similarity scores can be unreliable. 
However, use of abbreviations and a large number of related paragraphs for a topic make our observations particularly relevant for long form QA on technical documents. As future work, we would like to use RAG metrics \cite{es2023ragas, chen2023benchmarking} to choose retrieval strategies. Also, methods and evaluation metrics to answer follow-up questions would be of interest.

\newpage
\subsubsection*{URM Statement}
The authors acknowledge that at least one key author of this work meets the URM criteria of ICLR 2024 Tiny Papers Track.



\bibliography{iclr2023_conference_tinypaper}
\bibliographystyle{iclr2023_conference_tinypaper}

\newpage
\appendix

\section{Appendix A}\label{app:appA}
The prompts used for the LLM in our experiments are as follows:
\begin{itemize}
    \item System Prompt: Answer the questions based on the paragraphs provided here. DO NOT use any other information except that in the paragraphs. Keep the answers as short as possible. JUST GIVE THE ANSWER. NO PREAMBLE REQUIRED.
    \item User Prompt: ``PARAGRAPHS : ''+context + ``QUESTIONS: '' + query
\end{itemize}

\section{Appendix B}\label{app:appB}

\section{Appendix C - Supplementary Material}\label{app:appC}

We provide an \href{https://anonymous.4open.science/r/RAG_ICLR-55CB/README.md}{anonymized Git repository} which contains

\begin{itemize}
    \item Anonymized source code 
    \item Experiment v/s hypothesis tabulation (for consolidated quantitative results)
    \item Details of the experiments across 42 queries and 7 hypothesis
\end{itemize}

In addition, we provide details with respect to hypotheses in Table \ref{tab:observations} by providing sample queries and the retrieved and generated results. 


\section*{Supplementary material}

## Table 1 – Retrieval from Glossary

| ID | Query | Expected answer | Use full sentence | Use defined word | Use definition | Observations |
|---|---|---|---|---|---|---|
| 1. | Explain EIRP | effective isotropic radiated power (EIRP): The equivalent power of a transmitted signal in terms of an isotropic (omnidirectional) radiator. The EIRP equals the product of the transmitter power and the antenna gain (reduced by any coupling losses between the transmitter and antenna). | **Effective isotropic radiated power (EIRP): The equivalent power of a transmitted signal in terms of an isotropic (omnidirectional) radiator. The EIRP equals the product of the transmitter power and the antenna gain (reduced by any coupling losses between the transmitter and antenna). (0.614)** | **Effective isotropic radiated power (EIRP): The equivalent power of a transmitted signal in terms of an isotropic (omnidirectional) radiator. The EIRP equals the product of the transmitter power and the antenna gain (reduced by any coupling losses between the transmitter and antenna (0.478)** | **Effective isotropic radiated power (EIRP): The equivalent power of a transmitted signal in terms of an isotropic (omnidirectional) radiator. The EIRP equals the product of the transmitter power and the antenna gain (reduced by any coupling losses between the transmitter and antenna (0.669)** | The retriever obtains the correct definition by all 3. Also keyword at beginning of sentence is picked up with high similarity in full sentence mode. |
| | | | Emergency alert system (EAS): A U.S. national public warning system. (**0.372**) | Emergency alert system (EAS): A U.S. national public warning system. (**0.337**) | IEEE 802.1X authentication', ' Extensible Authentication Protocol (EAP) authentication transported by the IEEE 802.1X protocol. (**0.360**) | |
| | | | Received channel power indicator (RCPI): An indication of the total channel power (signal, noise, and interference) of a received frame measured on the channel and at the antenna connector used to receive the frame. (**0.362**) | Extended service area (ESA): The area within which members of an extended service set (ESS) can communicate. An ESA is larger than or equal to a basic service area (BSA) and might involve several basic service sets (BSSs) in overlapping, disjoint, or both configurations. (**0.322**) | Master session key (MSK): Keying material that is derived between the Extensible Authentication Protocol (EAP) peer and exported by the EAP method to the Authentication Server (AS) (**0.291**) | |
| 2 | Why do we need a beam steering matrix | beamforming steering matrix: A matrix determined using knowledge of the channel between a transmitter and an intended receiver that maps from space-time streams to transmit antennas with the goal of improving the signal power or signal-to-noise ratio (SNR) at the intended receiver | beamformee: A station (STA) that receives a physical layer (PHY) protocol data unit (PPDU) that was transmitted using a beamforming steering matrix. (**0.626**) | beamforming: A spatial filtering mechanism used at a transmitter to improve the received signal power or signal-to-noise ratio (SNR) at an intended receiver. Syn beam steering. (**0.745**) | beamformer: A station (STA) that transmits a physical layer (PHY) protocol data unit (PPDU) using a beamforming steering matrix. (**0.426**) | the top-1 does not return the correct answer using the defined word. The definition does not return it in top-3. The full definition returns it in 3rd position. However top sim in definition > correct answer in defined word. Similarly wrong 2 answers in the full definition have higher sim than correct answer in full sentence |
| | | | beamforming: A spatial filtering mechanism used at a transmitter to improve the received signal power or signal-to-noise ratio (SNR) at an intended receiver. Syn beam steering. (**0.458**) | **beamforming steering matrix: A matrix determined using knowledge of the channel between a transmitter and an intended receiver that maps from space-time streams to transmit antennas with the goal of improving the signal power or signal-to-noise ratio (SNR) at the intended receiver. (0.443)** | beamforming: A spatial filtering mechanism used at a transmitter to improve the received signal power or signal-to-noise ratio (SNR) at an intended receiver. Syn beam steering. (**0.367**) | |

| | | | | | | |
|---|---|---|---|---|---|---|
| | | | beamforming steering matrix: A matrix determined using knowledge of the channel between a transmitter and an intended receiver that maps from space-time streams to transmit antennas with the goal of improving the signal power or signal-to-noise ratio (SNR) at the intended receiver. (**0.398**) | directed frame: See individually addressed. (**0.309**) | unknown_definition_18:NOTE These uses include calculation of transmit steering, calculation of recommended modulation and coding scheme (MCS), and calculation of calibration parameters. (**0.359**) | |
| 3 | Which framework supports higher precision traffic under a condition of limited resources | multi-level precedence and preemption (MLPP): A framework used with admission control for the treatment of traffic streams based on precedence, which supports the preemption of an active traffic stream by a higher precedence traffic stream when resources are limited. Preemption is the act of forcibly removing a traffic stream in progress in order to free up resources for another higher precedence traffic stream. | traffic category (TC): A label for medium access control (MAC) service data units (MSDUs) that have a distinct user priority (UP), as viewed by higher layer entities, relative to other MSDUs provided for delivery over the same link. Traffic categories are meaningful only to MAC entities that support quality of service (QoS) within the MAC data service. These MAC entities determine the UP for MSDUs belonging to a particular traffic category using the priority value provided with those MSDUs at the MAC service access point (MAC SAP). (**0.456**) | traffic classification (TCLAS): The specification of one of several types of matching filter to classify protocol data units (PDUs) or medium access control (MAC) service data units (MSDUs) as belonging to a particular traffic stream (TS). Depending on the type of classification, the filter is applied within the MAC sublayer management entity (MLME), above the MAC, or within the MAC itself. (**0.535**) | admission control: An algorithm intended to prevent the violation of parameterized service commitments made by the network to admitted flows by controlling the admittance of a new flow into a resource constrained network. (**0.405**) | Only the definition can extract the correct answer but similarity for the correct answer via definition is lesser than even the 3rd result from the other methods |
| | | | traffic specification (TSPEC): The quality-of-service (QoS) characteristics of a data flow to and from a QoS station (STA). (**0.437**) | traffic specification (TSPEC): The quality-of-service (QoS) characteristics of a data flow to and from a QoS station (STA). (**0.489**) | multi-level precedence and preemption (MLPP): A framework used with admission control for the treatment of traffic streams based on precedence, which supports the preemption of an active traffic stream by a higher precedence traffic stream when resources are limited. Preemption is the act of forcibly removing a traffic stream in progress in order to free up resources for another higher precedence traffic stream.(**0.398**) | |
| | | | traffic stream (TS): A set of medium access control (MAC) service data units (MSDUs) to be delivered subject to the quality-of-service (QoS) parameter values provided to the MAC in a particular traffic specification (TSPEC). TSs are meaningful only to MAC entities that support QoS within the MAC data service. These MAC entities determine the TSPEC applicable for delivery of MSDUs belonging to a particular TS using the priority parameter provided with those MSDUs at the MAC service access point (MAC SAP). (**0.411**) | traffic stream (TS): A set of medium access control (MAC) service data units (MSDUs) to be delivered subject to the quality-of-service (QoS) parameter values provided to the MAC in a particular traffic specification (TSPEC). TSs are meaningful only to MAC entities that support QoS within the MAC data service. These MAC entities determine the TSPEC applicable for delivery of MSDUs belonging to a particular TS using the priority parameter provided with those MSDUs at the MAC service access point (MAC SAP). (**0.461**) | traffic specification (TSPEC): The quality-of-service (QoS) characteristics of a data flow to and from a QoS station (STA). (**0.396**) | |
| 4. | How is Ethertype protocol discrimination | medium access control (MAC) service tuple: The collection of a MAC service data unit (MSDU) along with the associated source address, | unknown_definition_2: NOTE See IETF RFC 3610. (0.434) | peer-to-peer traffic specification (PTP TSPEC): The quality-of-service (QoS) characteristics of a data flow between non-access point (non-AP) QoS stations (STAs). (**0.413**) | frame: A unit of data exchanged between peer protocol entities. (**0.418**) | Unable to identify this despite it being available as a keyword in the actual definition |

| # | Question | Definition | Match 1 | Match 2 | Match 3 | Notes |
|---|---|---|---|---|---|---|
| | (EPD) identified? | destination addresses, priority, drop eligibility, service class, optional set of service_access_point_identifiers, and optional indication of whether the supplied MSDU is in **Ethertype protocol discrimination (EPD)** or logical link control (LLC) protocol discrimination (LPD) format, which are all passed as parameters across the MAC service access point (SAP) and are all except the service_access_point_identifiers delivered across the distribution system between access points (APs), mesh gates, and the portal of an extended service set (ESS). | frame: A unit of data exchanged between peer protocol entities. (**0.432**) | protocol instance: An execution of a particular protocol that consists of the state of the communicating parties as well as the messages exchanged. (**0.380**) | unknown_definition_9: NOTE See IETF RFC 4282. (**0.407**) | |
| | | | unknown_definition_9: NOTE See IETF RFC 4282. (**0.404**) | traffic classification (TCLAS): The specification of one of several types of matching filter to classify protocol data units (PDUs) or medium access control (MAC) service data units (MSDUs) as belonging to a particular traffic stream (TS). Depending on the type of classification, the filter is applied within the MAC sublayer management entity (MLME), above the MAC, or within the MAC itself. (**0.370**) | traffic classification (TCLAS): The specification of one of several types of matching filter to classify protocol data units (PDUs) or medium access control (MAC) service data units (MSDUs) as belonging to a particular traffic stream (TS). Depending on the type of classification, the filter is applied within the MAC sublayer management entity (MLME), above the MAC, or within the MAC itself. (**0.383**) | |
| 5 | What does GAS stand for? | registered location query protocol (RLQP): The query protocol for registered location information that is received and transported by generic advertisement service (**GAS**) Public Action frames. | unknown_definition_1: NOTE See IETF RFC 2903 [B35]. (**0.384**) | unknown_definition_8: NOTE IEEE Std 802.11 supports only downlink (DL) MU-MIMO. See downlink multi-user multiple input, multiple output (DL-MU-MIMO) (in 3.2). (**0.343**) | distribution system medium (DSM): The medium or set of media used by a distribution system (DS) for communications between access points (APs), mesh gates, and the portal of an extended service set (ESS). (**0.357**) | Unable to identify this despite it being available as a keyword in the actual definition |
| | | | unknown_definition_2: NOTE See IETF RFC 3610. (**0.376**) | unknown_definition_13: NOTE For the purposes of this Standard, there is at most one portal in a given extended service set s (ESS s) infrastructure. In an implementation, a single logical portal function may be provided by multiple devices that provide integration services for the ESS. How such multiple devices coordinate to appear as a single logical portal is implementation dependent. (**0.337**) | service hash: A value used for representing a service. This value is formed from a hash of the service name. (**0.322**) | |
| | | | unknown_definition_9: NOTE See IETF RFC 4282. (**0.370**) | unknown_definition_18: NOTE These uses include calculation of transmit steering, calculation of recommended modulation and coding scheme (MCS), and calculation of calibration parameters. (**0.334**) | subscription service provider (SSP): An organization (operator) offering connection to network services, perhaps for a fee. (**0.304**) | |
| 6 | What is derived between EAP peer | master session key (MSK): Keying material that is derived between the Extensible Authentication Protocol (EAP) peer and exported by the EAP method to the Authentication Server (AS). | IEEE 802.1X authentication: Extensible Authentication Protocol (EAP) authentication transported by the IEEE 802.1X protocol. (**0.491**) | peer mesh station (STA): A mesh STA to which a mesh peering has been established. (**0.460**) | frame: A unit of data exchanged between peer protocol entities. (**0.625**) | The keyword is relatively at the beginning of the sentence and is therefore picked up wherever available (it is not available in the defined word) |
| | | | master session key (MSK): Keying material that is derived between the Extensible Authentication Protocol (EAP) peer and exported by the EAP method to the Authentication Server (AS). (**0.434**) | peer-to-peer link: A direct link within a quality-of-service (QoS) basic service set (BSS), a tunneled direct-link setup (TDLS) link, or a station-to-station (STA-to-STA) communication in an independent basic service set (IBSS). (**0.401**) | IEEE 802.1X authentication: Extensible Authentication Protocol (EAP) authentication transported by the IEEE 802.1X protocol. (**0.476**) | |
| | | | peer-to-peer traffic specification (PTP TSPEC): The quality-of-service (QoS) characteristics of a data flow between non-access point (non-AP) QoS stations (STAs). (**0.410**) | peer-to-peer traffic specification (PTP TSPEC): The quality-of-service (QoS) characteristics of a data flow between non-access point (non-AP) QoS stations (STAs). (**0.373**) | master session key (MSK): Keying material that is derived between the Extensible Authentication Protocol (EAP) peer and exported by the EAP method to the Authentication Server (AS). (**0.451**) | |

Table 2 – Retrieval from full document

| Query ID | Query | Expected answer | Search for paragraph | Search by sentence retrieve paragraph | Observations |
|---|---|---|---|---|---|
| F1 | What do the values of RAW Group Indication subfield in RPS element indicate? | The RAW Group Indication subfield indicates whether the RAW Group subfield is present in the RAW Assignment subfield and is interpreted as follows:<br>— When the RAW type is generic RAW, sounding RAW, or triggering frame RAW, the RAW Group Indication subfield indicates whether the RAW group defined in the current RAW assignment is the | The format of the RAW Group subfield is shown in Figure 9-672 | When the RAW is an AP PM RAW, the RAW Group Indication subfield equal to 0 indicates that the RAW group does not include any of the non-AP STAs, and the RAW Group subfield is not present. When the RAW Group Indication subfield is equal to 1, the RAW Group subfield is present. | It is clear that similarity search at sentence level and retrieval at paragraph level gives significantly better results. Since we retrieve 3 distinct paragraphs there is far more context available at the generator create good responses |
| | | same RAW group as defined in the previous RAW assignment. When the RAW Group Indication subfield is equal to 0, the RAW group defined in the current RAW assignment is the same as the RAW group defined in the previous RAW assignment and the RAW Group subfield is not present in this RAW assignment. When the RAW Group Indication subfield is equal to 1, the RAW Group subfield is present in this RAW assignment. The RAW Group Indication subfield in the first RAW assignment is set to 0 to indicate the RAW group in the first RAW assignment is the same as the range of AIDs in all the TIM bitmaps in the S1G Beacon frame.<br>— When the RAW is a non-TIM RAW, the RAW Group Indication subfield is set to 0 and the RAW Group subfield is not present.<br>— When the RAW is an AP PM RAW, the RAW Group Indication subfield equal to 0 indicates that the RAW group does not include any of the non-AP STAs, and the RAW Group subfield is not present.<br>When the RAW Group Indication subfield is equal to 1, the RAW Group subfield is present. | When the RAW is an AP PM RAW, the RAW Group Indication subfield equal to 0 indicates that the RAW group does not include any of the non-AP STAs, and the RAW Group subfield is not present. When the RAW Group Indication subfield is equal to 1, the RAW Group subfield is present. | The RAW Group Indication subfield indicates whether the RAW Group subfield is present in the RAW Assignment subfield and is interpreted as follows: | |
| | | | The RAW Group Indication subfield indicates whether the RAW Group subfield is present in the RAW Assignment subfield and is interpreted as follows: | When the RAW type is generic RAW, sounding RAW, or triggering frame RAW, the RAW Group Indication subfield indicates whether the RAW group defined in the current RAW assignment is the same RAW group as defined in the previous RAW assignment. When the RAW Group Indication subfield is equal to 0, the RAW group defined in the current RAW assignment is the same as the RAW group defined in the previous RAW assignment and the RAW Group subfield is not present in this RAW assignment. When the RAW Group Indication subfield is equal to 1, the RAW Group subfield is present in this RAW assignment. The RAW Group Indication subfield in the first RAW assignment is set to 0 to indicate the RAW group in the first RAW assignment is the same as the range of AIDs in all the TIM bitmaps in the S1G Beacon frame. | |

Table 3 – Generator Performance

| Query ID | Query | Expected answer | Generated response – similarity by paragraph | Generated Response – similarity by sentence, retrieve paragraph | Observations |
|---|---|---|---|---|---|
| F1 | What do the values of RAW Group Indication subfield in RPS element indicate? | The RAW Group Indication subfield indicates whether the RAW Group subfield is present in the RAW Assignment subfield and is interpreted as follows:<br>— When the RAW type is generic RAW, sounding RAW, or triggering frame RAW, the RAW Group Indication subfield indicates whether the RAW group defined in the current RAW assignment is the same RAW group as defined in the previous RAW assignment. When the RAW Group Indication subfield is equal to 0, the RAW group defined in the current RAW assignment is the same as the RAW group defined in the previous RAW assignment and the RAW Group subfield is not present in this RAW assignment. When the RAW Group Indication subfield is equal to 1, the RAW Group subfield is present in this RAW assignment. The RAW Group Indication subfield in the first RAW assignment is set to 0 to indicate the RAW group in the first RAW assignment is the same as the range of AIDs in all the TIM bitmaps in the S1G Beacon frame.<br>— When the RAW is a non-TIM RAW, the RAW Group Indication subfield is set to 0 and the RAW Group subfield is not present.<br>— When the RAW is an AP PM RAW, the RAW Group Indication subfield equal to 0 indicates that the RAW group does not include any of the non-AP STAs, and the RAW Group subfield is not present.<br>When the RAW Group Indication subfield is equal to 1, the RAW Group subfield is present. | 'According to the given paragraphs, the values of the RAW Group Indication subfield in the RPS (Radio Parameter Set) element indicate whether the RAW Group subfield is present or not in the RAW Assignment subfield. | According to the given paragraphs, the values of the RAW Group Indication subfield in the RPS (Radio Path Selection) element indicate whether the RAW Group subfield is present in the RAW Assignment subfield. If the value of RAW Group Indication subfield is equal to 0, it means that the RAW group defined in the current RAW assignment is the same as the RAW group defined in the previous RAW assignment, and the RAW Group subfield is not present in this RAW assignment. If the value of RAW Group Indication subfield is equal to 1, it means that the RAW Group subfield is present in this RAW assignment | A much broader context in general being available for the LLM because of multiple retrieved paragraphs helps it in create good long form question answering. |
| F2 | What is IGTK | integrity group temporal key (IGTK): A random value, assigned by the broadcast/multicast source station (STA), which is used to protect group addressed medium access control (MAC) management protocol data units (MMPDUs) from that source STA. | Sure! Based on the given paragraphs, the answer to your question is:\n\nIGTK stands for "integrity group temporal key" | According to the given paragraphs, the answer to question 1 is: IGTK stands for "integrity group temporal key" | Very often the generated response is a trivial answer giving the full form of an abbreviated technical term but not giving it's definition |
| F3 | What is idle power indicator | idle power indicator (IPI): A physical layer (PHY) indication of the total channel power (noise and interference) as measured in the channel at the receiving antenna connector while the station (STA) is idle, i.e., neither transmitting nor receiving a frame. | IPI - idle power indicator | IPI - idle power indicator | Very often a term is just abbreviated in a response. This can sometimes be reduced if a user a question like "Define Idle Power Indicator" but since this is left to the user can create issues |



\end{document}